\documentclass[10pt,twocolumn,letterpaper]{article}

\usepackage{iccv}
\usepackage{times}
\usepackage{epsfig}
\usepackage{xspace}
\usepackage{graphicx}
\usepackage{amsmath}
\usepackage{amssymb}

\usepackage{booktabs}
\usepackage{tabulary}
\usepackage[dvipsnames]{xcolor}
\usepackage[caption=false]{subfig}
\usepackage[british,english,american]{babel}
\usepackage{soul}
\usepackage{xspace}\xspace
\usepackage{adjustbox}
\usepackage{array}

\usepackage{dblfloatfix}
\usepackage{transparent}
\usepackage{stackengine}
\usepackage{multirow}
\usepackage{algorithm}
\usepackage{listings}

\definecolor{citecolor}{HTML}{0071bc}
\usepackage[pagebackref=false,breaklinks=true,letterpaper=true,colorlinks,citecolor=citecolor,bookmarks=false]{hyperref}

\newcommand{\app}{\raise.17ex\hbox{$\scriptstyle\sim$}}

\newcommand{\apbbox}[1]{AP$^\text{bb}_\text{#1}$}
\newcommand{\apmask}[1]{AP$^\text{mk}_\text{#1}$}

\newcommand{\apdp}[1]{AP$^\text{gpsm}_\text{#1}$}

\newcolumntype{x}[1]{>{\centering\arraybackslash}p{#1pt}}
\newcolumntype{y}[1]{>{\raggedright\arraybackslash}p{#1pt}}
\newcolumntype{z}[1]{>{\raggedleft\arraybackslash}p{#1pt}}
\newlength\savewidth\newcommand\shline{\noalign{\global\savewidth\arrayrulewidth
  \global\arrayrulewidth 1pt}\hline\noalign{\global\arrayrulewidth\savewidth}}
\newcommand{\tablestyle}[2]{\setlength{\tabcolsep}{#1}\renewcommand{\arraystretch}{#2}\centering\footnotesize}

\makeatletter\renewcommand\paragraph{\@startsection{paragraph}{4}{\z@}
  {.5em \@plus1ex \@minus.2ex}{-.5em}{\normalfont\normalsize\bfseries}}\makeatother

\def\x{\times}

\usepackage{textpos}  % for superposing a text box

\iccvfinalcopy % *** Uncomment this line for the final submission

\def\iccvPaperID{9531}

\ificcvfinal\fi

\begin{document}

%%%%%%%%% TITLE
\title{Region Similarity Representation Learning}

\author{
\vspace{.5em}
 Tete Xiao$^{1}$\thanks{: Equal contribution.} \quad Colorado J Reed$^{1*}$ \quad Xiaolong Wang$^2$ \quad Kurt Keutzer$^1$ \quad Trevor Darrell$^1$ \vspace{.5em}\\
 $^1$UC Berkeley\quad$^2$UC San Diego \vspace{.3em}
}

\maketitle
% Remove page # from the first page of camera-ready.
\ificcvfinal\thispagestyle{empty}\fi

\newcommand{\ours}{ReSim\xspace}

%%%%%%%%% ABSTRACT
\begin{abstract}
We present Region Similarity Representation Learning (\ours), a new approach to self-supervised representation learning for localization-based tasks such as object detection and segmentation.
While existing work has largely focused on solely learning global representations for an entire image, 
\ours learns both regional representations for localization as well as semantic image-level representations.
\ours operates by sliding a fixed-sized window across the overlapping area between two views (\eg, image crops), aligning these areas with their corresponding convolutional feature map regions, and then maximizing the feature similarity across views.
As a result, \ours learns spatially and semantically consistent feature representation throughout the convolutional feature maps of a neural network. 
A shift or scale of an image region, \eg, a shift or scale of an object, has a corresponding change in the feature maps; this allows downstream tasks to leverage these representations for localization.
Through object detection, instance segmentation, and dense pose estimation experiments, we illustrate how \ours learns representations which significantly improve the localization and classification performance compared to a competitive MoCo-v2 baseline: $+2.7$ AP$^{\text{bb}}_{75}$ VOC, $+1.1$ AP$^{\text{bb}}_{75}$ COCO, and $+1.9$ AP$^{\text{mk}}$ Cityscapes. Code and pre-trained models are released at: \url{https://github.com/Tete-Xiao/ReSim} 

\end{abstract}

%%%%%%%%% BODY TEXT
\section{Introduction}

\begin{figure}[t] 
    \centering
    \includegraphics[width=\linewidth]{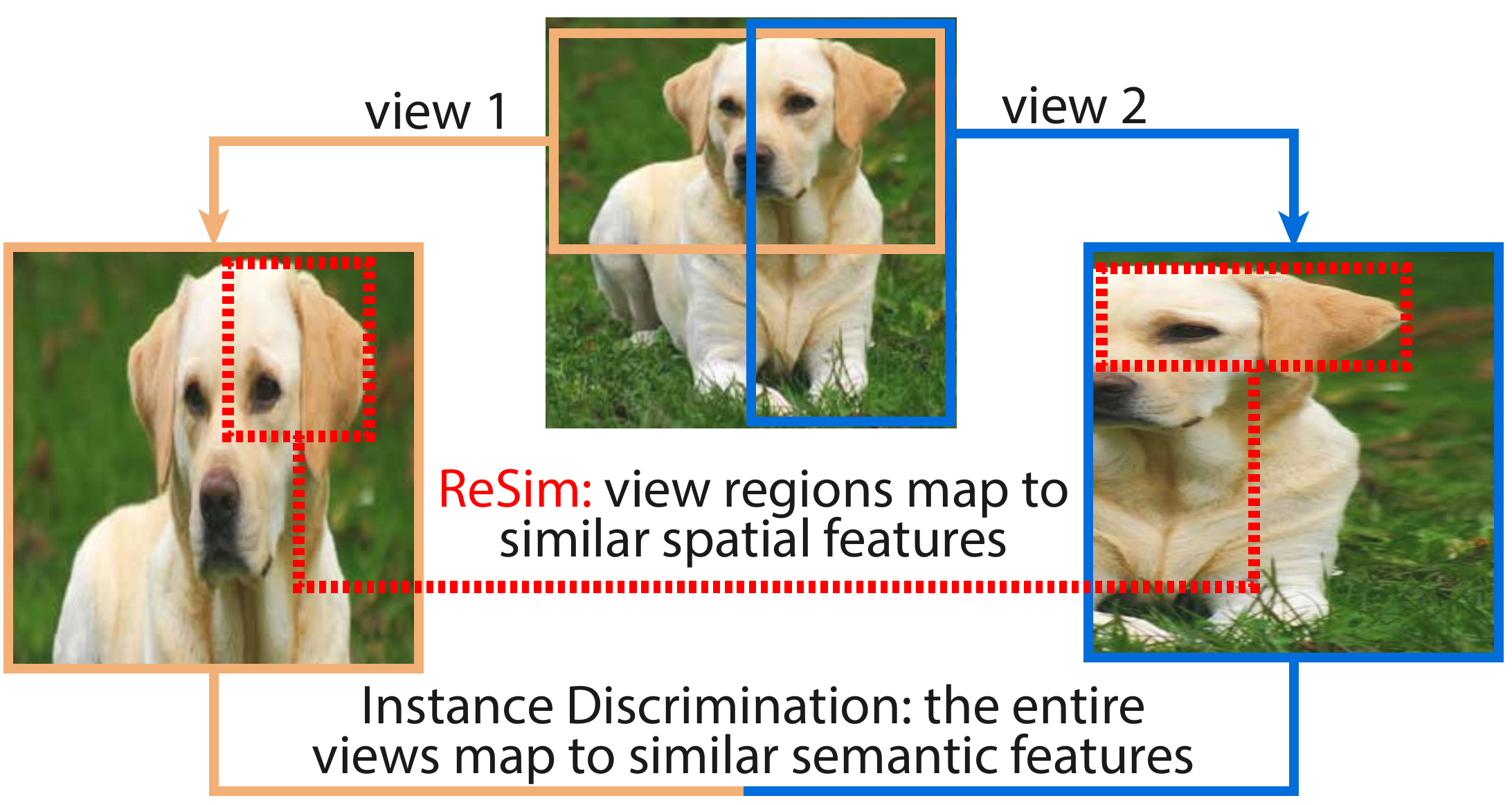}
     \caption{Existing instance discrimination-based self-supervised learning frameworks learn representations by augmenting an image into two different views (\eg, cropping/scaling the input image) and then maximizing the similarity between the image features for the entire views. In this work, we present Region Similarity Representation Learning (\ours), which learns representations by maximizing the similarity of corresponding sub-image \emph{regions} throughout the convolutional layers of a network.
     In the above example, instance discrimination learns to map both views to the same features, despite the fact that the dog's eyes and ears are in different locations. On the other hand, \ours learns features which explicitly align these changes with corresponding changes in the convolutional feature maps.
     }
     \label{fig:resim-teaser}
\end{figure}

\begin{figure*}[t] 
    \centering
    \includegraphics[width=0.9\linewidth]{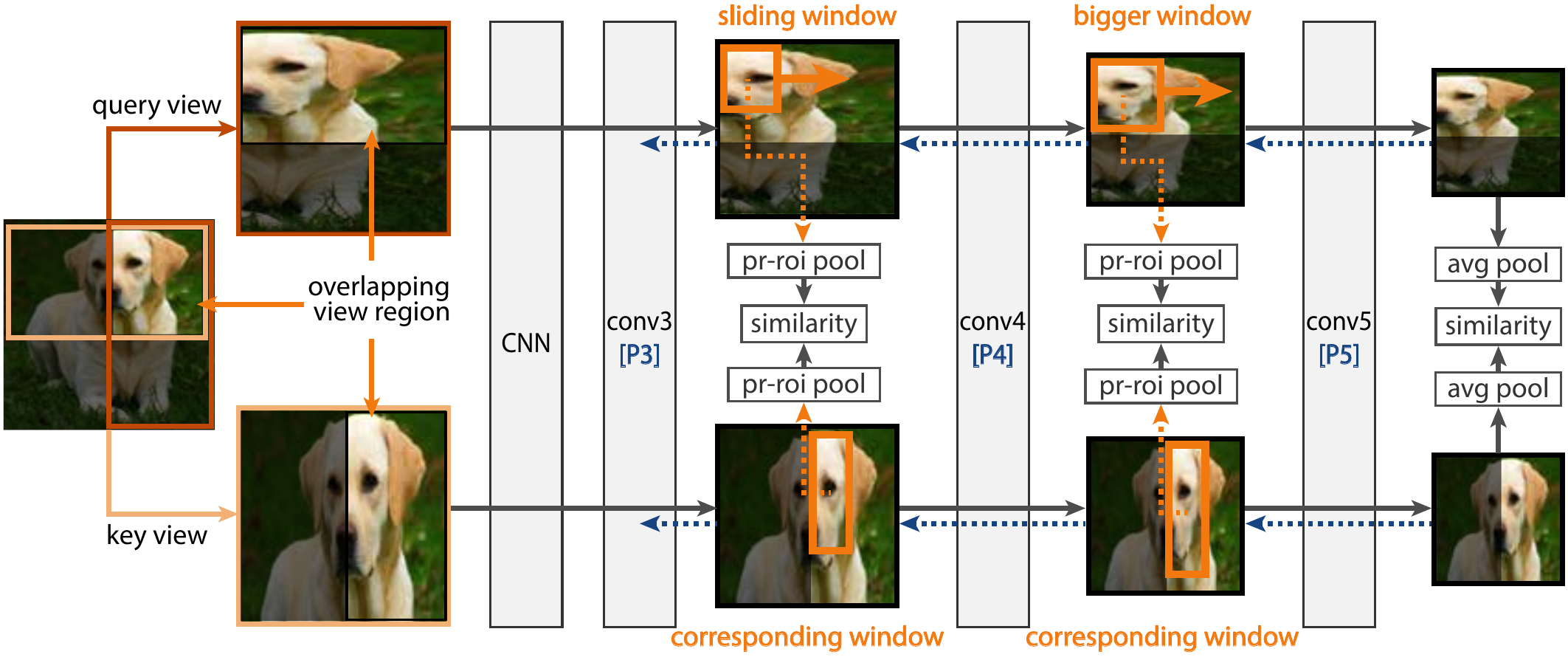}
     \caption{\ours takes two different views of an image as input, \ie, a \emph{query} and \emph{key} view obtained by cropping/scaling an image. The views have an associated overlapping area as highlighted in each of the above views. Both views are encoded using the same network (CNN), \eg, a ResNet-50~\cite{he2016deep}. Before the final convolutional layers in the network, \ours slides a fixed-size window over the overlapping area between the two views, aligns window regions with corresponding regions in the convolutional feature maps using Precise RoI Pooling from~\cite{jiang2018acquisition}, and then maximizes the similarity between these features. Earlier layers use smaller sliding windows as the feature maps have higher spatial resolution. Furthermore, similar to Feature Pyramid Networks~\cite{lin2017feature}, the feature maps are combined with semantic top-down features from later convolution layers, as indicated by the blue arrows and ``[P3]/[P4]/[P5]'' layers. The feature maps following the final convolutional layer are used for instance discrimination learning following either~\cite{chen2020improved} or~\cite{chen2020exploring}. 
     }
     \label{fig:resim-pipeline}
\end{figure*}

% Introduce instance discrimination learning and the problem of invariances 
Recently, self-supervised pre-training has outperformed supervised pre-training for a number of computer vision applications such as image classification and object detection \cite{caron2020unsupervised, chen2020simple, chen2020improved}. 
Much of this recent progress comes from exploiting the \emph{instance discrimination} task~\cite{bachman2019learning, dosovitskiy2014discriminative, malisiewicz2008recognition,wu2018unsupervised,ye2019unsupervised}, in which a network learns image-level features that are invariant to certain image augmentations. Specifically, instance discrimination maximizes the similarity of two \emph{views} of an image, where each view is an augmented version of an image, while minimizing its similarity to views which originate from other images~\cite{caron2020unsupervised, chen2020simple, chen2020exploring}. 

Chen et al.~\cite{chen2020simple} compared a diverse set of possible augmentations, and found that random cropping and scaling have the largest impact on downstream ImageNet~\cite{deng2009imagenet} classification performance. Several follow-up works have further explored augmentation policies and confirmed this finding~\cite{reed2020selfaugment,tian2020makes,xiao2020should}. Through cropping and scaling augmentations and similarity maximization, the network learns to map various scales and crops of an image to the same feature representations. For example, an image crop of the top half of a dog's body and the right half of a dog's body would map to the same representation in the embedding space, which a downstream task could then classify as ``dog''. 

However, instance discrimination uses global image-level feature representations for these views, which is obtained by average pooling the final convolutional feature map. It does not enforce any type of spatial consistency in the convolutional features (see Figure~\ref{fig:resim-teaser}, ``instance discrimination'' path). For example, different crops that scale and shift the dog's ear will not necessarily have a corresponding scale and shift in the convolutional feature maps throughout the network -- instance discrimination only optimizes the final globally pooled features. This is problematic for downstream tasks such as object detection that leverage the spatial information from the convolutional feature maps for object localization.

To address this issue, we introduce Region Similarity Representation Learning (\ours): a self-supervised pre-training method which learns spatially consistent features across multiple convolutional layers.
Inspired by the Region Proposal Network (RPN) used in Faster-RCNN~\cite{ren2016faster}, \ours operates by sliding a fixed-size window across the overlapping region between two image views, mapping the corresponding regions in each view to their associated regions in the convolutions layers throughout the network, and then maximizing the similarity of these convolution feature regions, along with the global similarity objective. 
%By learning representations on regions of convolutional feature maps and global representations from the entire image, \ours can be interpreted as simultaneously learning localization sensitive and location invariant representations at different layers. Taken together, the goal of these embedding spaces is to eliminate the inductive bias introduced by maximizing the global similarity of two randomly cropped and scaled views~\cite{xiao2020should}.
See Figure~\ref{fig:resim-teaser} for a high-level difference between \ours and existing instance discrimination techniques, and see Figure~\ref{fig:resim-pipeline} for a detailed description of the full \ours pipeline.

As we show, maximizing the similarity of these convolutional feature map regions leads to representations that improve object localization for downstream detection and instance segmentation tasks. Furthermore, we extend the framework to learn features at various scales by using sliding windows of multiple sizes at different feature maps. We adopt the Feature Pyramid Network (FPN) design from Lin et al.~\cite{lin2017feature}, a design which naturally incorporates feature hierarchies and propagates stronger semantic features to earlier convolutional layers through the top-down path. Region-level self-supervised similarity learning trains feature pyramid layers without labeled supervision and leads to further improvement on downstream tasks.

We conduct object detection, instance segmentation, and dense pose estimation experiments on PASCAL VOC~\cite{everingham2007pascal}, COCO~\cite{lin2014microsoft}, and Cityscapes~\cite{cordts2016cityscapes} and show that \ours learns representations which significantly improve the classification and localization performance compared to a MoCo-v2 baseline, \ie, $+2.7$ AP$^\text{bb}_{75}$ on VOC, $+1.1$ AP$^{\text{bb}}_{75}$ on COCO, and $+1.9$ AP$^{\text{mk}}$ on Cityscapes.

\section{Related work}
\label{sec:relwork}

\paragraph{Self-supervised representation learning.}
The goal of representation learning is to reveal the intrinsic qualities of data in such a way that they are informative and effective for a desired task~\cite{bengio2013representation}.
Practically, this often manifests as pre-training a deep network so that it can be finetuned for a particular downstream task, see~\cite{devlin2018bert, donahue2014decaf, erhan2010does,goodfellow2016deep, he2019momentum, henaff2019data, lecun2015deep, radford2018improving, zeiler2014visualizing}. Recently, SimCLR~\cite{chen2020simple} and MoCo~\cite{chen2020improved, he2019momentum} demonstrated substantial improvements by using similar forms of instance contrastive learning where a network was trained to identify a pair of views originating from the same image when contrasted with a large set of views from other images.
Following SimCLR and MoCo, later works, such as SwAV~\cite{caron2020unsupervised} and BYOL~\cite{grill2020bootstrap}, reported substantial improvements for image classification tasks, but as several follow-up works have shown~\cite{ding2021unsupervised, xie2021detco, yang2021instance}, SwAV and BYOL do not tend to lead to improvements on localization-based tasks such as object detection and segmentation. This decrease in performance indicated that strictly optimizing global, image-level representations could decrease performance for tasks which require localization.

In contrast to prior work that leveraged instance discrimination to learn global, image-level representations, we propose a region-based pretext task to learn representations for tasks which require both localization and semantic classification. In earlier work, several authors proposed representation learning at the pixel or region level via color prediction~\cite{vondrick2018tracking, zhang2016colorful}, key-point encoding \cite{thewlis2017unsupervised, minderer2019unsupervised}, optical flow similarity~\cite{dosovitskiy2015flownet}, and cycle consistency or frame prediction in videos~\cite{li2019joint, wang2019learning}. Our work differs in that we build on instance discrimination pretext task and simultaneously learn semantic features from an entire image as well as regional features across multiple scales, shifts, and resolutions.

\paragraph{Object detection and instance segmentation.}
Object detection and segmentation localize and classify object bounding boxes or pixels within an image -- see \cite{liu2020deep} for a survey and review. Many commonly used object detection and instance segmentation techniques such as Fast-RCNN~\cite{girshick2015fast}, Faster-RCNN~\cite{ren2016faster}, and Mask-RCNN~\cite{he2017mask} operate through a two-stage process in which they extract region features from convolutional feature maps, regress the location of the region to align with a ground truth bounding box, and then classify the region. 

When regressing the location of the regional features, small adjustments to the region location within the convolutional feature map can have a large impact on the classification, \eg, shifting a region may cause it to overlap with different objects in the input pixel space. 
As a result, the regional features of the convolutional feature maps require spatial sensitivity for regression as well as semantic sensitivity for classification. Existing self-supervised methods have largely focused on learning strong semantic representations by average pooling the final convolutional layer, and they do not explicitly maintain spatial consistency throughout the convolutional layers. In this work, we propose a self-supervised learning method which learns both spatially consistent and semantically sensitive regional features.

\paragraph{Pixel and grid-based contrastive learning.}
Recently, several related papers emerged: \cite{pinheiro2020unsupervised} explored pixel-level contrastive learning, whereas our work examines region-level consistency across the convolutional features. Wang et al.~ \cite{wang2020dense} proposed contrastive learning using a grid of feature vectors over the feature maps. Rather than maximizing the similarity of the regions, their solution was more akin to clustering, in that they compared all combination of feature vectors from the feature grid and maximized the similarity of the most similar pair.
As we show below, our method demonstrates superior performance over these models.

\section{Region Similarity Representation Learning}
This section presents \ours: a self-supervised pre-training method which learns spatially consistent representations on feature pyramids.
\begin{figure}[t] 
    \centering
    \includegraphics[width=0.9\linewidth]{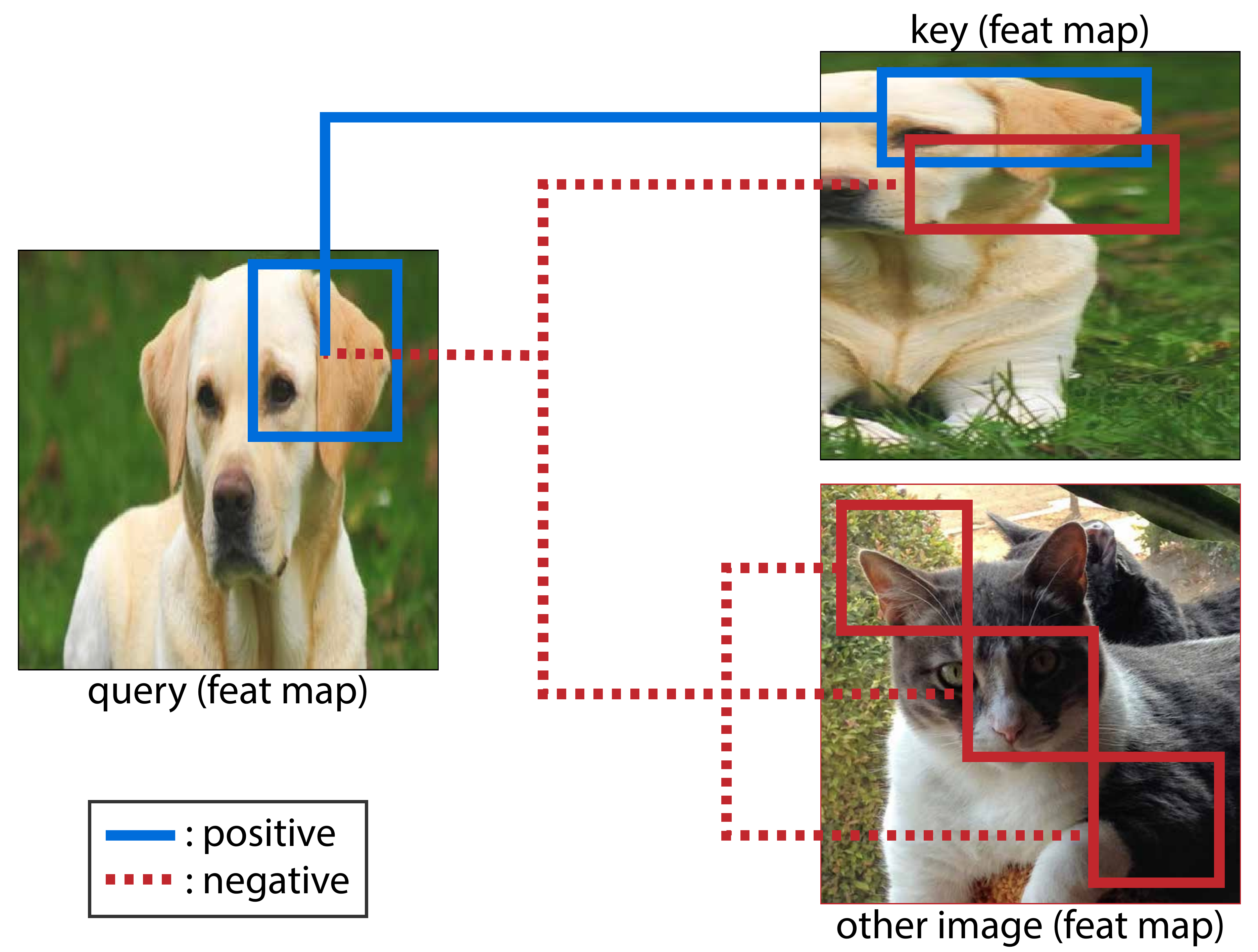}
     \caption{At each of the final convolutional layers, \ours maximizes the similarity of aligned convolutional feature map regions across the query and key views of an image -- the \emph{positive} samples. \ours simultaneously minimizes the similarity with a set of \emph{negative} samples: the non-positive regions from the same key view (potentially overlapping with the positive region) and any other region from other images.
     }
     \vspace{-1.0mm}
     \label{fig:resim-pos-neg}
\end{figure}
\subsection{Preliminaries}
% Region and representation definition; % Our Objective

State-of-the-art self-supervised representation learning frameworks overwhelmingly exploit instance discrimination as their pretext task, see~\cite{jaiswal2021survey}. In instance discrimination, a reference image is augmented by a set of predefined augmentation modules $\mathcal{T}=[\mathcal{T}_1, \mathcal{T}_2, \cdots, \mathcal{T}_n]$, yielding two views: query and key. For instance, Figure~\ref{fig:resim-pipeline} shows the query and key view after cropping and scaling augmentations. The features from the final convolutional layer from the two views are average-pooled into image features that are subsequently enforced to be similar in the embedding space by a learning objective, such as contrastive loss~\cite{chen2020simple} or cosine similarity~\cite{chen2020exploring}.

In the context of representation learning, an ideal encoder for localization-sensitive tasks should consistently encode the same image region, \eg, the same object components across different views of an image, to the same point in the representational embedding space.
This is necessary for tasks such as object detection with Faster-RCNN~\cite{ren2016faster} or related detectors~\cite{he2017mask,liu2020deep}, which localize and then classify regions within the convolutional feature maps. When regressing the location of the regional features, small adjustments to the region within the convolutional feature map can have a large impact on the object classification prediction at the pixel level, \eg, shifting a feature map region by a few pixels may cause the region to overlap with different objects in the input pixel space, changing the ground-truth classification and regression target.

Similarly, different image regions, \eg, different object components across two different views of an image, should map to different points in the embedding space. By aligning similar image regions to the same representations and dissimilar image regions to dissimilar representations, downstream tasks can leverage the representations to localize and classify the underlying image components. As illustrated in Figure~\ref{fig:resim-pipeline}, the key and query views of the dog image correspond to the same image even though the two views are scaled and translated differently. 

When learning representations for image classification, existing works map all regions of an image to similar representations~\cite{chen2020simple, chen2020improved}. This occurs because the widely adopted set of augmentations from Chen et al.~\cite{chen2020simple} include Random Resized Crop (RRC), which enforces that the query and key views always correspond to different regions in the original image.  
Therefore, we propose a region-level similarity learning component (Figure~\ref{fig:resim-pipeline}) which enforces that the same regions within the views encode to the same spatially and semantically consistent feature representations.

\subsection{Framework Overview}
The full \ours framework, shown in Figure~\ref{fig:resim-pipeline}, operates as follows: First, \ours augments an input image into two different views -- a query and key, where each view is passed through a common encoder, \eg, a ResNet-50. On a subset of the final convolutional layers, \eg, \texttt{conv3} and \texttt{conv4}, \ours slides a fixed-size window over the query view, but only within the overlapping region between the query and key view -- this area is highlighted in Figure~\ref{fig:resim-pipeline}. Following a foundational idea from feature pyramids~\cite{lin2017feature}, earlier layers use a smaller window because they have a higher spatial resolution. \ours then uses Precise RoI Pooling~\cite{jiang2018acquisition} to extract a feature vector from the associated feature map region for both views. 

\ours uses a similarity optimization function to maximize the similarity of the aligned feature vectors from the corresponding view regions. We chose a contrastive optimization function from~\cite{he2019momentum}, which also takes dissimilar regions from the same image as well as other images as negative examples, as illustrated in Figure~\ref{fig:resim-pos-neg}. Note: we also explored a cosine similarity optimization function from \cite{chen2020exploring} which does not require negative examples -- see \S\ref{sec:exp}.

Finally, following the final convolutional layer, \ours adopts the global average pool of the final feature map for an instance-level contrastive optimization from \cite{he2019momentum} or a cosine similarity optimization from \cite{chen2020exploring}. As shown by the blue dotted arrows in Figure~\ref{fig:resim-pipeline}, \ours uses 1x1 convolutional layers to propagate semantically strong features to the earlier convolutional layers; these are known as \emph{top-down} connections in the original FPN work~\cite{lin2017feature}.

\subsection{Region-level Similarity}
We formulate a region as a rectangular sub-area of an image $\mathcal{I}$ described by its top, left, bottom, and right coordinates $(t, l, b, r)$. An encoder $f$ yields the representation of any region of a given image, \ie, $f\left(\mathcal{I}, \left(t, l, b, r\right)\right)$. Figure~\ref{fig:resim-pipeline} shows a CNN encoder taking two image regions as input.
Our objective is to perform self-supervised pre-training of the encoder in such a way that the learned representations are tailored for downstream tasks such as object detection and instance segmentation.

\paragraph{Aligning corresponding regions.}
As shown in Figure~\ref{fig:resim-pipeline}, \ours performs self-supervised similarity learning on sub-image level regions across two views of an image. 
Given two augmented views $\mathcal{I}_q$ and $\mathcal{I}_k$ from the same input image $\mathcal{I}$, and a region in the query view denoted by its coordinates $(t_q, l_q, b_q, r_q)$, we need to find its corresponding region in the key view. 
From the set of widely adopted augmentations listed in the previous subsection, the only spatial transformations which affect the image coordinates are RRC and horizontal flips. We ensure that the query and key views are \emph{either both horizontally flipped or both not flipped}, as otherwise they should have difference ground-truth regression term for downstream tasks (see object regression in~\cite{girshick2015fast}). Thus, it leaves us with RRC alone. Denote the operator as $\mathcal{R}$, the problem can be formulated as, given $\mathcal{I}_q = \mathcal{R}_q(\mathcal{I})$, $\mathcal{I}_k = \mathcal{R}_k(\mathcal{I})$ and $(t_q, l_q, b_q, r_q)$, find $(t_k, l_k, b_k, r_k) = \mathcal{R}_{q\rightarrow k}(t_q, l_q, b_q, r_q)$. The solution is trivial as RRC is simple linear translations, therefore $\mathcal{R}_{q\rightarrow k} = \mathcal{R}_k(\mathcal{R}^{-1}_q)$, where $\mathcal{R}^{-1}$ is the inverse translation of the applied spatial transformations.

\paragraph{Generating candidate regions and extracting features.}
\ours generates candidate regions for the region-level contrastive learning by sliding a small window across the overlapping region between the query and key view -- see Figure~\ref{fig:resim-pipeline}. The overlapping areas between the query and key views are selected as valid areas for the sliding window because they can be mapped between the views. 

Rather than cropping regions from query/key images and feeding them into encoders, we encode an entire image and extract region features through Precise RoI Pooling~\cite{jiang2018acquisition} which takes in a rectangular window and a convolutional feature map, and yields features of the input window in the size of $h{\times}w{\times}c$, where we set $h=1$ and $w=1$ to create a feature vector from the region, and $c$ is the same as the number of input feature map channels. Given a candidate region on the input views, \ours performs region-level similarity across multiple convolutional feature maps at the end of the network, e.g., using $C3$ and $C4$ as shown in Figure~\ref{fig:resim-pipeline}.
 
\paragraph{Region similarity on feature pyramids.}
Lin et al.~\cite{lin2017feature} proposed Feature Pyramid Networks (FPNs), an object detection architecture which combined the low-resolution, semantically strong features from later convolutional layers with high-resolution, semantically weak features from earlier layers via \emph{top-down connections}. 
As shown by the blue dotted arrows in Figure~\ref{fig:resim-pipeline}, we add top-down connections to propagate semantically strong features to the earlier convolutional layers used for region similarity learning --- we call this variant \ours-FPN. We follow the proposed methodology from~\cite{lin2017feature}: we use lateral convolutional connections to map the output of the convolutional layers to FPN layers (indicated with $\{P3, P4, P5\}$ in Figure~\ref{fig:resim-pipeline}), and then directly add the top-down layers after using nearest neighbor sampling to match the spatial resolutions and $1{\times1}$ convolutional layers to match the channel dimensions.

\paragraph{Similarity learning objectives.}
Denote the 4-tuple of a region $(t, l, b, r)$ as $u$, the set of valid region pairs in query and key from image $\mathcal{I}$ as $\{(u_q^1, u_k^1), (u_q^2, u_k^2), \dots, (u_q^n, u_k^n) \}$, and the region-level similarity function
\begin{equation}
    E_{i,j}^{\{k_+, k_-\}} = \exp{\left(f(\mathcal{I}_q, u_q^i) \cdot f(\mathcal{I}_{\{k_+, k_-\}}, u_{\{k_+, k_-\}}^j) / \tau\right)},
\end{equation}
where the positive sample $\mathcal{I}_q = \mathcal{T}_q(I)$, $\mathcal{I}_k = \mathcal{T}_k(I)$, $f(I, u)$ is the Precise RoI-pooled feature of region $u$ from image $\mathcal{I}$, and the negative pairs can be \emph{any} image and region pairs not identical to $\mathcal{I}_q$ and $u_q$ -- see Figure~\ref{fig:resim-pos-neg}. $\tau$ is a temperature parameter to regularize the similarity distribution.
We learn the region-level similarity for a query via the following objective:
\begin{equation}
    \mathcal{L}_q^{\text{rs}} = \frac{1}{n}\sum_{i=1}^{n}{\frac{E_{i,i}^{k_+}}{E_{i,i}^{k_+} + \sum_{j \neq i}E_{i, j}^{k+} + \sum_{k_-,j}{E_{i, j}^{k_-}}}},
\end{equation}

Denote the image-level similarity function with global average pooling (GAP) as
\begin{equation}
    D^{\{k_+, k_-\}} = \exp{\left(f(\mathcal{I}_q, \text{GAP}) \cdot f(\mathcal{I}_{k\{ +, -\}}, \text{GAP}) / \tau\right)}.
\end{equation}
We use the global image-level similarity objective from~\cite{he2019momentum}:
\begin{equation}
    \mathcal{L}_q^{\text{is}} = \frac{D^{k_+}}{D^{k_+} + \sum_{k_-}D^{k_-}}.
\end{equation}

We combine the region-level similarity and global image-level similarity objective via a weighting hyperparameter, $\lambda$,  to yield the final objective:
\begin{equation}
    \mathcal{L}_q = \mathcal{L}_q^{\text{rs}} + \lambda\mathcal{L}_q^{\text{is}}.
\end{equation}

\subsection{Framework Configurations}
\label{sec:method:details}
We use two $3{\times}3$ convolutions on $C4/P4$ or $P3$ to project the feature maps to 128 channels. The first convolution is followed by Batch Normalization and ReLU activation.
The similarity boxes shown in Figure~\ref{fig:resim-pipeline} compute the similarity between aligned convolutional feature map regions. For this similarity computation, we use a momentum contrastive similarity with the associated settings from~\cite{chen2020improved} as default, while we also investigated a cosine similarity with settings from~\cite{chen2020exploring} -- see \S\ref{sec:ablation}.

For the region-based contrastive similarity, we take negative samples to be both the non-positive regions from the same key view and any other region from other images, see Figure~\ref{fig:resim-pos-neg}. While MoCo uses a momentum-based queue to maintain a large number of negative image-level samples, we find that such a queue is unnecessary for region-level samples as there are a large number of negative region samples within each batch, \ie, we can generate over 30,000 negatives within a mini-batch of 256 images on $C4$ -- see the appendix for more details. 

We design two variations of \ours: 1) \ours-C4, which applies region-similarity learning on ResNet $C4$ feature map \emph{without} FPN; and 2) \ours-FPN, which applies region-similarity learning on FPN $P3$ and $P4$ feature maps.
When applying the Precise RoI Pooling operator to obtain the features for the convolutional feature map regions, we apply the pooling operator on $C4$/$P4$ with a down-sampling rate of 16, and on feature map $P3$ with a down-sampling rate of 8. By default, we use a sliding window of 48 pixels with a stride of 32 pixels on the input image on $C4$/$P4$, and a sliding window of 32 pixels with a stride of 24 pixels on $P3$. We ablate these settings in \S~\ref{sec:ablation}.

% size and stride of the windows

\definecolor{Gray}{gray}{0.5}

\newcommand{\randinit}{\tablestyle{1pt}{1} \begin{tabular}{z{21}y{26}} \multicolumn{2}{c}{\demph{random init.}} \end{tabular}}
% ------------------------------------------------

\newcommand{\demph}[1]{\textcolor{Gray}{#1}}
\newcommand{\std}[1]{{\fontsize{5pt}{1em}\selectfont ~~$_\pm$$_{\text{#1}}$}}

\definecolor{Highlight}{HTML}{39b54a}  % green

\renewcommand{\hl}[1]{\textcolor{Highlight}{#1}}

\newcommand{\res}[3]{
\tablestyle{1pt}{1}
\begin{tabular}{z{16}y{18}}
{#1} &
\fontsize{7.5pt}{1em}\selectfont{~(${#2}${#3})}
\end{tabular}}

\newcommand{\reshl}[3]{
\tablestyle{1pt}{1} 
\begin{tabular}{z{16}y{18}}
{#1} &
\fontsize{7.5pt}{1em}\selectfont{~\hl{(${#2}$\textbf{#3})}}
\end{tabular}}

\newcommand{\resrand}[2]{\tablestyle{1pt}{1} \begin{tabular}{z{16}y{18}} \demph{#1} & {} \end{tabular}}
\newcommand{\ressup}[2]{\tablestyle{1pt}{1} \begin{tabular}{z{16}y{18}} {#1} & {} \end{tabular}}

\begin{table}[tb]
\centering
\tablestyle{1pt}{1.0}
\begin{tabular}{x{56}|x{54}|x{54}x{54}c}
pre-train &
AP &
AP$_\text{50}$ &
AP$_\text{75}$ & \\ 
\shline
\randinit & \resrand{33.8}{} & \resrand{60.2}{} & \resrand{33.1}{} & \\
supervised & \ressup{44.0}{} & \ressup{72.8}{} & \ressup{45.5}{} & \\
MoCo-v2 & \ressup{54.4}{} & \ressup{80.1}{} & \ressup{60.0}{} & \\
\hline
\ours-C4 & \reshl{\textbf{55.9}}{+}{1.5} & \reshl{\textbf{81.3}}{+}{1.2} & \reshl{\textbf{62.1}}{+}{2.1} & \\
\end{tabular}
\vspace{.3em}
% ------------------------------------------------
\caption{\textbf{Comparisons of \ours and MoCo-v2~\cite{he2019momentum} on PASCAL VOC object detection task.} The models are pre-trained on \emph{IN-100}, the weights of which are transferred to a Faster R-CNN R50-C4 subsequently finetuned on VOC~\texttt{trainval07+12}, and evaluated on~\texttt{test2007}. AP$_\text{k}$ is the average precision at $k$ IoU threshold, and AP is the COCO-style metric which averages scores from \texttt{[0.5:0.95:0.05]}. In the brackets are the deltas to the MoCo-v2 baseline.}
\label{tab:baseline_on_voc}
\end{table}

\section{Experiments}
\label{sec:exp}
We study the transfer capability of \ours for localization-dependent downstream tasks. We perform \emph{self-supervised pre-training} on two splits of the ImageNet dataset~\cite{deng2009imagenet}: 1) 1000-category ImageNet (IN-1K), the standard ImageNet training set containing ${\sim}$1.25M images; and 2) 100-category ImageNet (IN-100), a subset of IN-1K split containing ${\sim}$125k images, following previous works~\cite{tian2019contrastive,xiao2020should}. The pre-trained model is subsequently finetuned on PASCAL VOC~\cite{everingham2010pascal} for object detection, COCO~\cite{lin2014microsoft} for instance segmentation, Cityscapes~\cite{cordts2016cityscapes} for instance segmentation, and DensePose~\cite{guler2018densepose} for dense pose estimation. For the ablation studies, we use IN-100 for pre-training and report full IN-1K pre-trained results for selected best models.

\paragraph{Training.}
Our hyperparameters closely follow the adopted self-supervised learning framework MoCo-v2~\cite{chen2020improved}. We use a ResNet-50~\cite{he2016deep} backbone, pre-training for 200 epochs for IN-1K, 500 epochs for IN-100, with a mini-batch size of 256 on 8 GPUs. The initial learning rate is set as 0.03 and follows a cosine decaying schedule~\cite{loshchilov2016sgdr}. The weight decay is 0.0001, SGD momentum is 0.9, and the augmentations are the same as~\cite{chen2020improved}. We use 1.0 for loss balancing term $\lambda$. We use a momentum queue of 65,536 for IN-1K experiments, and of 16,384 for IN-100 experiments. \ours-FPN takes ${\sim}$62 hours to train 200 IN1K-epochs with 8 NVIDIA V100 GPUs, compared to ${\sim}53$ hours for MoCo. Note that the design of \ours does \emph{not} affect the complexity of its transferred downstream model.

\subsection{IN-100 Study}
\label{sec:ablation}
We first experiment with various \ours implementation decisions using IN-100, and from these results, we then study the performance of \ours on IN-1k.
\paragraph{Setup.}
Throughout these experiments, we adopt the commonly used PASCAL VOC object detection task~\cite{everingham2015pascal}. We use a Faster R-CNN~\cite{ren2016faster} with R50-C4~\cite{he2017mask} backbone. As in~\cite{he2019momentum}, we unfreeze all Batch Normalization layers and synchronizing their statistics across GPUs~\cite{peng2018megdet} during training. The short-side length of input images is randomly selected from [480, 800] pixels during training and fixed at 800 for inference. Training is performed on \verb|trainval07+12| set (${\sim}$16.5k images), and testing is performed on \verb|test2007| set (${\sim}$4.9k images), unless otherwise specified. The network is trained on 8 GPUs of mini-batch size 16. Training takes 24,000 iterations with a base learning rate 0.02. The learning rate is decreased by a factor of $1/10$ at the 18,000 and 22,000 iteration. All settings follow~\cite{he2019momentum}. 

\begin{table}[tb]
\small
\centering
\tablestyle{1pt}{1.0}
\begin{tabular}{x{56}|x{54}|x{54}x{54}c}
pre-train &
AP &
AP$_\text{50}$ &
AP$_\text{75}$ & \\ 
\shline
SimSiam & \ressup{54.6}{} & \ressup{80.6}{} & \ressup{60.7}{} & \\
+\ours-C4 & \reshl{55.7}{+}{1.1} & \reshl{81.2}{+}{0.6} & \reshl{61.9}{+}{1.2} & \\
\hline
MoCo-v2 & \ressup{54.4}{} & \ressup{80.1}{} & \ressup{60.0}{} & \\
+\ours-C4 & \reshl{\textbf{55.9}}{+}{1.5} & \reshl{\textbf{81.3}}{+}{1.2} & \reshl{\textbf{62.1}}{+}{2.1} & \\
\end{tabular}
\vspace{.3em}
% ------------------------------------------------
\caption{\textbf{Building \ours on various self-supervised representation learning frameworks}, \ie, SimSiam~\cite{chen2020exploring} and MoCo-v2. All models are obtained by self-supervised pre-training on IN-100, transferring weights to a Faster R-CNN R50-C4 detector subsequently finetuned on VOC~\texttt{trainval07+12}, and evaluated on~\texttt{test2007}. We use the prediction and projection heads as in~\cite{chen2020exploring} for \ours on SimSiam. The improvement over SimSiam shows that \ours is applicable to multiple frameworks.}
\label{tab:simsiam_on_voc}
\end{table}

%#########

\paragraph{Comparisons with baselines.}
We first compare the \ours framework with similarity learning on C4 (\ours-C4) to the MoCo-v2 baseline. Since \ours is built on MoCo-v2, the key difference between the two frameworks is the presence of similarity learning across the convolutional feature map regions in \ours. 
Table~\ref{tab:baseline_on_voc} shows the VOC transfer results of the two frameworks, as well as the models which are supervise pre-trained on ImageNet and randomly initialized. Albeit competing against a strong reference, \ours surpasses MoCo-v2 by a large margin of 1.5, 1.2 and 2.1 at AP, AP$_\text{50}$, and AP$_\text{75}$, respectively. The performance gap at high AP metric (AP$_\text{75}$) is larger than it at low AP metric (AP$_\text{50}$) which is less localization-dependent. 

\begin{table}[tb]
\small
\centering
% ------------------------------------------------
\subfloat[Full finetuning setting on VOC]{
\tablestyle{1pt}{1.0}
\begin{tabular}{x{45}|x{40}|x{35}|x{35}x{35}}
pre-train &
window &
AP &
AP$_\text{50}$ &
AP$_\text{75}$ \\ 
\shline
MoCo-v2 & N/A & 54.4 & 80.1 & 60.0 \\
\hline
\ours-C4 & W16-S16 & 55.0 & 80.7 & 61.3 \\
\ours-C4 & W48-S16 & 55.8 & 81.2 & 61.6 \\
\ours-C4 & W48-S32 & 55.9 & \textbf{81.3} & 62.1 \\
\ours-C4 & W64-S48 & \textbf{56.0} & 81.0 & \textbf{62.7} \\
\end{tabular}	
} % end of subfloat
% ------------------------------------------------
\\
\vspace{-.5em}
% ------------------------------------------------
\subfloat[Frozen-backbone setting, finetuning on VOC]{
\tablestyle{1pt}{1.0}
\begin{tabular}{x{45}|x{40}|x{35}|x{35}x{35}}
pre-train &
window &
AP &
AP$_\text{50}$ &
AP$_\text{75}$ \\ 
\shline
MoCo-v2 & N/A & 47.1 & 75.4 & 50.6 \\
\hline
\ours-C4 & W16-S16 &  48.7 & 76.0 & 53.2 \\
\ours-C4 & W48-S16 &  49.5 & 76.7 & 54.1 \\
\ours-C4 & W48-S32  & \textbf{50.0} & \textbf{77.2} & \textbf{55.0} \\
\ours-C4 & W64-S48  & 49.7 & \textbf{77.2} & 54.3 \\
\end{tabular}	
} % end of subfloat
% ------------------------------------------------
\vspace{.3em}
\caption{\textbf{Comparisons of various sliding window sizes and strides under (a) full-finetuning and (b) frozen-backbone settings.}
``W$l$-S$k$'' denotes sliding a window of size $l{\times}l$ at stride $k$. \ours-C4 by default uses W48-S32 setting.
Inspired by the linear classification metric for classification, in the frozen-backbone setting the backbone of object detector is frozen from finetuning to directly evaluate the quality of features for object detection.
The more significant improvement in the frozen-backbone setting implies \ours learns features of better quality for localization-dependent task.
}
\vspace{-1.0mm}
\label{tab:rs_frozen_voc}
\end{table}

\paragraph{Self-supervised learning frameworks.}
We investigated \ours with an additional self-supervised learning framework. Specifically, we selected the recently proposed Siamese-based framework, SimSiam~\cite{chen2020exploring}, for its simplicity and effectiveness. We use identical designs of its projection and prediction heads (see their reference for details) for \ours as the image-level similarity learning heads presented in SimSiam for consistency. Note that since those heads have several sequentially connected 2048 channel layers rather than a single 2048-128 channel connection used in MoCo-v2, the design significantly increases computational complexity over \ours on MoCo, \ie, \ours-FPN based on SimSiam takes 46.7 hours to train on IN-100, compared to 15.5 hrs for \ours-FPN based on MoCo.

Table~\ref{tab:simsiam_on_voc} shows the results. \ours built upon SimSiam improves the baseline by 1.1, 0.6 and 1.2 at AP, AP$_\text{50}$, and AP$_\text{75}$, respectively, and is comparable to \ours built upon MoCo. Due to the extra pre-training time, we choose MoCo as our backbone framework.

\paragraph{Frozen backbone for finetuning on VOC.}
Linear classification, \ie, training a linear classifier on a frozen backbone is widely adopted to evaluate the representations \emph{for classification}~\cite{caron2020unsupervised, chen2020simple, chen2020exploring, he2019momentum, grill2020bootstrap, wu2018unsupervised}. We design a similar feature evaluation experiment on object detection. After transferring the self-supervise pre-trained weights to a Faster R-CNN object detector, we freeze the backbone and finetune the RPN and object classifier head on labeled object data, as RPN is randomly initialized and object classifier is task-specific. For the Faster R-CNN R50-C4 detector, specifically, all weight layers from $C1$ to $C4$ feature map, along with statistics of Batch Normalization layers are frozen; weights in RPN and C5 (object classification head) are finetuned.  According to Goyal et al.~\cite{goyal2019scaling}, this type of evaluation is ideal for representation learning because finetuning an entire network ``evaluates not only the quality of the representations but also the initialization and optimization method.''

Table~\ref{tab:rs_frozen_voc}(b) shows the results. \ours improves the baseline MoCo-v2 (default W48-S32) by 2.9, 1.8, 4.4 at AP, AP$_\text{50}$, and AP$_\text{75}$, respectively, a much larger margin compared to the full-finetune setting. The improvement is particularly significant at high IoU metric AP$_\text{75}$. It indicates \ours yields features of higher quality for localization-dependent tasks, particularly when limiting the impact of the initialization and optimizations used during finetuning.

%---------------- All VOC
\begin{table}[tb]
\small
\centering
\tablestyle{1pt}{1.0}
\begin{tabular}{x{56}|x{54}|x{54}x{54}c}
pre-train &
AP &
AP$_\text{50}$ &
AP$_\text{75}$ & \\ 
\shline
\randinit & \resrand{33.8}{} & \resrand{60.2}{} & \resrand{33.1}{} & \\
supervised & \ressup{53.5}{} & \ressup{81.3}{} & \ressup{58.8}{} & \\\hline
Jigsaw~\cite{goyal2019scaling} & \res{48.9}{-}{4.6} & \res{75.1}{-}{6.2} & \res{52.9}{-}{5.9} & \\
Rotation~\cite{goyal2019scaling} & \res{46.3}{-}{7.2} & \res{72.5}{-}{8.8} & \res{49.3}{-}{9.5} & \\
NPID++~\cite{misra2020self} & \res{52.3}{-}{1.2} & \res{79.1}{-}{2.2} & \res{56.9}{-}{1.9} & \\
SimCLR~\cite{chen2020simple} & \res{51.5}{-}{2.0} & \res{79.4}{-}{1.9} & \res{55.6}{-}{3.2} & \\
PIRL~\cite{misra2020self} & \res{54.0}{+}{0.5} & \res{80.7}{-}{0.6} & \res{59.7}{+}{0.9} & \\
BoWNet~\cite{gidaris2020learning} & \reshl{55.8}{+}{2.3} & \res{81.3}{+}{0.0} & \reshl{61.1}{+}{2.3} & \\
MoCo~\cite{he2019momentum} & \reshl{55.9}{+}{2.4} & \res{81.5}{+}{0.2} & \reshl{62.6}{+}{3.8} & \\
MoCo-v2~\cite{chen2020improved} & \reshl{57.0}{+}{3.5} & \reshl{82.4}{+}{1.1} & \reshl{63.6}{+}{4.8} & \\
SwAV~\cite{caron2020unsupervised} & \reshl{56.1}{+}{2.6} & \reshl{82.6}{+}{1.3} & \reshl{62.7}{+}{3.9} & \\
DenseCL~\cite{wang2020dense} & \reshl{58.7}{+}{5.2} & \reshl{82.8}{+}{1.5} & \reshl{65.2}{+}{6.4} & \\\hline
\ours-C4 & \reshl{58.7}{+}{5.2} & \reshl{\textbf{83.1}}{+}{1.8} & \reshl{\textbf{66.3}}{+}{7.5} & \\
\ours-FPN & \reshl{\textbf{59.2}}{+}{5.7} & \reshl{82.9}{+}{1.6} & \reshl{65.9}{+}{7.1} & \\
\end{tabular}
% ------------------------------------------------
\caption{\textbf{Comparison with previous works on PASCAL VOC}~\texttt{trainval07+12}, \textbf{evaluated on}~\texttt{test2007}. Jigsaw and Rotation results are from~\cite{misra2020self}. SimCLR result is from~\cite{wang2020dense}. Both \ours-C4 and \ours-FPN improve MoCo-v2 baseline, and achieve state-of-the-art over competitive concurrent work DenseCL~\cite{wang2020dense}. In the brackets are the deltas to the ImageNet supervised pre-training baseline. \textcolor{Highlight}{\textbf{Green}}: deltas at least +1.0.}
\vspace{-2.5mm}
\label{tab:all_voc}
\end{table}%----------------

\renewcommand{\res}[3]{
\tablestyle{1pt}{1}
\begin{tabular}{z{14}y{18}}
{#1} &
\fontsize{7.5pt}{1em}\selectfont{(${#2}${#3})}
\end{tabular}}

\renewcommand{\reshl}[3]{
\tablestyle{1pt}{1}
\begin{tabular}{z{14}y{18}}
{#1} &
\fontsize{7.5pt}{1em}\selectfont{\textcolor{Highlight}{(${#2}$\textbf{#3})}} 
\end{tabular}}

\renewcommand{\resrand}[2]{\tablestyle{1pt}{1} \begin{tabular}{z{14}y{18}} \demph{#1} & {} \end{tabular}}
\renewcommand{\ressup}[2]{\tablestyle{1pt}{1} \begin{tabular}{z{14}y{18}} {#1} & {} \end{tabular}}

\newcommand{\cgap}[2]{
\fontsize{6pt}{1em}\selectfont{(${#1}${#2})}
}
\newcommand{\cgaphl}[2]{
\fontsize{6pt}{1em}\selectfont{\textcolor{Highlight}{(${#1}$\textbf{#2})}}
}

\renewcommand{\randinit}{\tablestyle{1pt}{1} \begin{tabular}{z{21}y{26}} \multicolumn{2}{c}{random init.} \end{tabular}}

\begin{table*}[ht]
	\small
	\resizebox{1.0\linewidth}{!}{
		\subfloat[Mask R-CNN R50-FPN, 1$\x$ schedule]{
			\tablestyle{.8pt}{1.05}
			\begin{tabular}{cr|
					z{17}y{18}
					z{17}y{18}
					z{17}y{18}|
					z{17}y{18}
					z{17}y{18}
					z{17}y{18}c
				}
				pretrain & ~ &
				\multicolumn{2}{c}{\apbbox{~}} &
				\multicolumn{2}{c}{\apbbox{50}} &
				\multicolumn{2}{c|}{\apbbox{75}} &
				\multicolumn{2}{c}{\apmask{~}} &
				\multicolumn{2}{c}{\apmask{50}} &
				\multicolumn{2}{c}{\apmask{75}} &\\
				\shline
				\demph{random init} & ~ &
				\demph{31.0} & ~ & \demph{49.5} & ~ & \demph{33.2} & ~ &
				\demph{28.5} & ~ & \demph{46.8} & ~ & \demph{30.4} & ~ & \\
				supervised & ~ &
				38.9 & ~ & 59.6 & ~ & 42.7 & ~ &
				35.4 & ~ & 56.5 & ~ & 38.1 & ~ & \\					

				MoCo-v2 & ~ &
				38.9 & \cgap{+}{0.0} & 59.2 & \cgap{-}{0.4} & 42.4 & \cgap{-}{0.3} &
				35.4 & \cgap{+}{0.0} & 56.2 & \cgap{-}{0.3} & 37.8 & \cgap{-}{0.3} \\	
				
				VADeR~\cite{pinheiro2020unsupervised} & ~ &
				39.2 & \cgap{+}{0.3} & 59.7 & \cgap{+}{0.1} & 42.7 & \cgap{+}{0.0} &
				35.6 & \cgap{+}{0.2} & 56.7 & \cgap{+}{0.2} & 38.2 & \cgap{+}{0.1} \\	
				
				DenseCL~\cite{wang2020dense}$^\dagger$ & ~ &
				39.4 & \cgaphl{+}{0.5} & 59.9 & \cgap{+}{0.3} & 42.7 & \cgap{+}{0.0} &
				35.6 & \cgap{+}{0.2} & 56.7 & \cgap{+}{0.2} & 38.2 & \cgap{+}{0.1} \\	
				\hline	
				ReSim-C4 & ~ &
				39.3 & \cgap{+}{0.4} & 59.7 & \cgap{+}{0.1} & 43.1 & \cgap{+}{0.4} & 35.7 & \cgap{+}{0.3} & 56.7 & \cgap{+}{0.2} & 38.1 & \cgap{+}{0.0} & \\
				ReSim-FPN & ~ &
				39.5 & \cgaphl{+}{0.6} & 59.9 & \cgap{+}{0.3} & 43.3 & \cgaphl{+}{0.6} &
				35.8 & \cgap{+}{0.4} & 57.0 & \cgaphl{+}{0.5} & 38.4 & \cgap{+}{0.3} \\

				ReSim-FPN$^T$ & ~ &
				\textbf{39.8} & \cgaphl{+}{0.9} & \textbf{60.2} & \cgaphl{+}{0.6} & \textbf{43.5} & \cgaphl{+}{0.8} &
				\textbf{36.0} & \cgaphl{+}{0.6} & \textbf{57.1} & \cgaphl{+}{0.6} & \textbf{38.6} & \cgaphl{+}{0.5} \\
				\hline
				ReSim-FPN$^T$ (400 ep) & ~ &
				\textbf{40.3} & \cgaphl{+}{1.4} & \textbf{60.6} & \cgaphl{+}{1.0} & \textbf{44.2} & \cgaphl{+}{1.5} &
				\textbf{36.4} & \cgaphl{+}{1.0} & \textbf{57.5} & \cgaphl{+}{1.0} & \textbf{38.9} & \cgaphl{+}{0.8} \\
			\end{tabular}	
		}  
		\subfloat[Mask R-CNN R50-FPN, 2$\x$ schedule]{
		\tablestyle{.8pt}{1.05}
			\begin{tabular}{
					z{17}y{18}
					z{17}y{18}
					z{17}y{18}|
					z{17}y{18}
					z{17}y{18}
					z{17}y{18}c
				}
				\multicolumn{2}{c}{\apbbox{~}} &
				\multicolumn{2}{c}{\apbbox{50}} &
				\multicolumn{2}{c|}{\apbbox{75}} &
				\multicolumn{2}{c}{\apmask{~}} &
				\multicolumn{2}{c}{\apmask{50}} &
				\multicolumn{2}{c}{\apmask{75}} &\\
				\shline
				%\demph{\randinit} &
				\demph{36.7} & ~ & \demph{56.7} & ~ & \demph{40.0} & ~ & 
				\demph{33.7} & ~ & \demph{53.8} & ~ & \demph{35.9} & ~ &\\					
				%\supimgnet &
				40.6 & ~ & 61.3 & ~ & 44.4 & ~ &
				36.8 & ~ & 58.1 & ~ & 39.5 & ~ & \\

				%MoCo v2 & ~ &
				40.9 & \cgap{+}{0.3} & 61.5 & \cgap{+}{0.2} & 44.6 & \cgap{+}{0.2} &
				37.0 & \cgap{+}{0.2} & 58.4 & \cgap{+}{0.3} & 39.6 & \cgap{+}{0.1} \\
				
				%VADeR & ~ &
				- &  & - &   & - &   &
				- &  & - &   & - &   & \\

				%DenseCL & ~ &
				41.2 & \cgaphl{+}{0.6} & 61.9 & \cgaphl{+}{0.6} & 45.1 & \cgaphl{+}{0.7} &
				37.3 & \cgaphl{+}{0.5} & 58.9 & \cgaphl{+}{0.8} & 40.1 & \cgaphl{+}{0.6} \\\hline
				% ReSim-C4
				41.1 & \cgaphl{+}{0.5} & 61.5 & \cgap{+}{0.2} & 44.8 & \cgap{+}{0.4} &
				37.1 & \cgap{+}{0.3} & 
				58.6 & \cgaphl{+}{0.5} & 
				39.8 & \cgap{+}{0.3} \\
				
				% ReSim-FPN
				\textbf{41.4} & \cgaphl{+}{0.8} & 61.8 & \cgaphl{+}{0.5} & 45.4 & \cgaphl{+}{1.0} &
				\textbf{37.5} & \cgaphl{+}{0.7} & \textbf{59.1} & \cgaphl{+}{1.0} & \textbf{40.4} & \cgaphl{+}{0.9} \\
				
				% ReSim-FPN-T
				\textbf{41.4} & \cgaphl{+}{0.8} & \textbf{61.9} & \cgaphl{+}{0.6} & \textbf{45.4}  & \cgaphl{+}{1.0} &
				\textbf{37.5} & \cgaphl{+}{0.7} & \textbf{59.1} & \cgaphl{+}{1.0} & 40.3 & \cgaphl{+}{0.8} \\
				\hline
				% ReSim-FPN-T
				\textbf{41.9} & \cgaphl{+}{1.3} & \textbf{62.4} & \cgaphl{+}{1.1} & \textbf{45.9}  & \cgaphl{+}{1.5} &
				\textbf{37.9} & \cgaphl{+}{1.1} & \textbf{59.4} & \cgaphl{+}{1.3} & \textbf{40.6} & \cgaphl{+}{1.1} \\				
			\end{tabular}	
		}  
	} 
    \vspace{0.1mm}
	\caption{\textbf{Results on COCO object detection and instance segmentation tasks.} The models are pre-trained on IN-1K for 200 epochs (except the final entry, which shows extended performance out to 400 epochs), the weights of which are transferred to Mask R-CNN R50-FPN model, subsequently finetuned on~\texttt{train2017} (1$\x$ and 2$\x$ schedules) and evaluated on~\texttt{val2017}. Averaging precision on bounding-boxes (AP$^{\text{bb}}$) and masks (AP$^{\text{mk}}$) are used as benchmark metrics. All \ours models outperform VADeR~\cite{pinheiro2020unsupervised} and the MoCo-v2 counterpart, surpassing the supervised pre-training under all metrics (we report VADeR results from their paper which did not include a 2x schedule). In the brackets are the gaps to the ImageNet supervised pre-training baseline. \textcolor{Highlight}{\textbf{Green}}: deltas at least +0.5. $^\dagger$: Pre-trained weights are downloaded from the official releases of the authors and trained on our COCO frameworks, as~\cite{wang2020dense} adopted a different COCO finetuning baseline settings from the common approach used in this paper and other previous works~\cite{he2019momentum}.
	}\vspace{-3.0mm}
	\label{tab:coco}
\end{table*}

\paragraph{Region size and stride.}
We vary the size and stride of sliding windows in \ours-C4 to determine the optimal combination. The results are reported in Table~\ref{tab:rs_frozen_voc}(a,b). ``W$l$-S$k$'' denotes sliding a window of size $l{\times}l$ at stride $k$. By considering the performance both at frozen-backbone and standard full-finetune settings, we choose W48-S32 setting as the default for \ours-C4.

\paragraph{RoI Align vs. Precise RoI Pooling.}
We compare Precise RoI Pooling~\cite{jiang2018acquisition} and RoI Align~\cite{he2017mask} as the region feature extraction operator. On VOC, \ours using RoI Align achieves AP/AP$_\text{50}$/AP$_\text{75}$ of 55.7/81.0/62.0 with all parameters finetuned, and 48.9/76.6/53.0 with a frozen backbone. Comparing these results with Table~\ref{tab:rs_frozen_voc} indicates that both RoI Align and Precise RoI Pooling work well, but in the frozen-backbone setting, RoI Align performs worse than Precise RoI Pooling. We conjecture this is because Precise RoI Pooling does not heuristically set the number of sampling points, enabling a more robust and simpler optimization.

\begin{figure}[t] 
    \centering
    \includegraphics[width=0.9\linewidth]{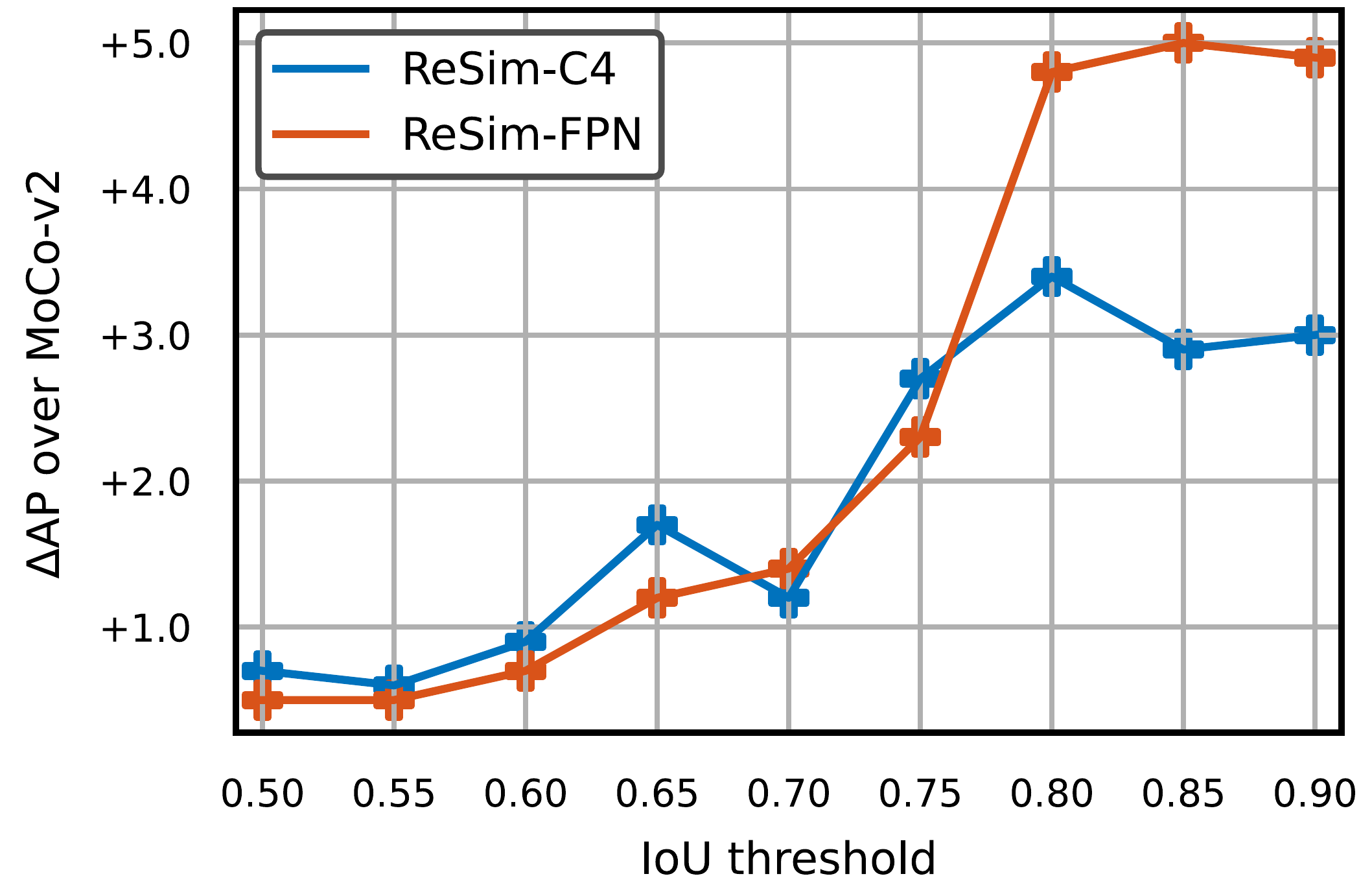}
     \caption{\textbf{$\Delta$AP of \ours over MoCo-v2 at various IoU thresholds on PASCAL VOC.} As the IoU threshold increases, both \ours methods perform considerably better than MoCo-v2, especially at higher IoU threshold, indicating that the \ours features have better localization capability.
     }
     \label{fig:iou_curve}
     \vspace{-1.0mm}
\end{figure}

\subsection{IN-1K Results}
Following the IN-100 experiments, we studied pre-training on the full IN-1K and report the results on PASCAL VOC, Cityscapes, COCO detection and COCO dense pose estimation. We evaluate two models as introduced in \S\ref{sec:method:details}: 1) \ours-C4, which performs region-level similarity on $C4$ feature map of ResNet; and 2) \ours-FPN, which performs region-level similarity on the $P3$ and $P4$ feature maps on the top-down path of FPN. Only the backbone ResNet weights are transferred for finetuning downstream tasks. On the other hand, \emph{if the network of a downstream task adopts the FPN design}, then we can transfer the FPN weights from \ours-FPN to the new network. We term this as \ours-FPN$^T$, and report the results on selected downstream tasks which use FPN as its baseline.

\paragraph{PASCAL VOC.} We use the Faster R-CNN R50-C4 detector on VOC and adopt the same setup for PASCAL VOC as described in \S\ref{sec:ablation}. We first report the results of finetuning on ~\texttt{trainval07+12} and evaluating on~\texttt{test2007}. 
Table~\ref{tab:all_voc} compares our method with a series of previous state-of-the-arts. Respectively, \ours-C4 and \ours-FPN improve over their baseline MoCo-v2 by $1.7/0.7/2.7$ and $2.2/0.5/2.3$ at AP/AP$_\text{50}$/AP$_\text{75}$, and claims state-of-the-art results over the competitive concurrent work DenseCL~\cite{wang2020dense}. 

In Figure~\ref{fig:iou_curve} we show the AP improvements of \ours over MoCo-v2 at IoU threshold from 0.5 to 0.9. Not only do both \ours methods outperform MoCo-v2, the larger improvement at high IoU indicates superior localization accuracy, particular for \ours-FPN. The growing performance gap as the IoU threshold increases indicates that \ours has led to improved localization of the objects.

We also report the results of finetuning on PASCAL VOC~\texttt{trainval2007} and evaluating on~\texttt{test2007}. \ours-C4 achieves $48.9/75.4/53.5$ at AP/AP$_\text{50}$/AP$_\text{75}$, in comparison with $46.6/72.8/50.9$ for MoCo-v2.

\paragraph{COCO Object Detection and Instance Segmentation.} 
We use Mask-RCNN~\cite{he2017mask} with R50-FPN~\cite{lin2017feature} as our base model and add new Batch Normalization layers before the FPN parameters. All Batch Normalization statistics are synchronized across GPUs. The short-side length of input images is randomly selected from [640, 800] pixels during training and fixed at 800 for inference. Training is performed on~\texttt{train2017} split with ${\sim}$118k images, and testing is performed on~\texttt{val2017} split. Two metrics are used, Average Precision on bounding-boxes (AP$^{\text{bb}}$), and Average Precision on masks (AP$^{\text{mk}}$). We report finetuning results on the standard 1x and 2x schedules, the latter of which trains twice as long as the former.

Table~\ref{tab:coco} shows the results. All variations of \ours improve over its MoCo-v2 counterpart under both box and mask metrics, outperforming the supervised pre-training baseline. Unlike MoCo-v2, \ours improves with the 1$\x$ schedule. We note that \ours-FPN outperforms \ours-C4, and in turn \ours-FPN$^T$, in which FPN weights are transferred, outperforms \ours-FPN, demonstrates the effectiveness of our framework on feature pyramids. Furthermore, the final row in Table~\ref{tab:coco} shows a continued increase in performance after 400 epochs of pre-training, indicating that \ours continues to its representations in longer pre-training regimes.

\begin{table}[t]
\centering

%% coco dp
\begin{raggedleft}
\tablestyle{2pt}{1.1}
\begin{tabular}{x{56}|x{48}x{48}x{48} c}
\multicolumn{1}{c|}{} &
\multicolumn{3}{c}{\fontsize{7.5pt}{1em}\selectfont \textbf{COCO dense pose estimation}} & 
\\
pre-train & 
\apdp{~} & \apdp{50} & \apdp{75} &  % coco pose
\\
\shline
supervised
& \ressup{65.8}{} & \ressup{92.6}{} & \ressup{77.8}{}  % coco pose
& \\
MoCo-v2
& \res{66.2}{+}{0.4} & \reshl{93.1}{+}{0.5} & \res{77.4}{-}{0.4}  % coco pose
& \\\hline
\ours-C4
& \reshl{66.6}{+}{0.8} & \res{92.9}{+}{0.3} & \reshl{78.5}{+}{0.7}  % coco pose
& \\
\ours-FPN
& \reshl{\textbf{67.1}}{+}{1.3} & \reshl{\textbf{93.2}}{+}{0.6} & \reshl{\textbf{80.0}}{+}{2.2}  % coco pose
& \\
\ours-FPN$^T$
& \reshl{66.8}{+}{1.0} & \res{92.8}{+}{0.2} & \reshl{79.2}{+}{1.4}  % coco pose
& \\
\end{tabular}
\end{raggedleft}
\vspace{.3em}

%% coco dp
\begin{raggedleft}
\tablestyle{2pt}{1.1}
\begin{tabular}{x{56}|x{64}x{64} c}
\multicolumn{1}{c|}{} &
\multicolumn{2}{c}{\fontsize{7.5pt}{1em}\selectfont \textbf{Cityscapes Instance Seg.}} & 
\\
pre-train & 
\apmask{~} & \apmask{50} &  % coco pose
\\
\shline
supervised
& \ressup{33.0}{} & \ressup{59.8}{}   % coco pose
& \\
MoCo-v2
& \reshl{33.7}{+}{0.7} & \reshl{61.8}{+}{2.0}  % coco pose
& \\\hline
\ours-C4
& \reshl{35.4}{+}{2.4} & \reshl{63.1}{+}{3.3}   % coco pose
& \\
\ours-FPN
& \reshl{\textbf{35.6}}{+}{2.6} & \reshl{\textbf{63.4}}{+}{3.6}  % coco pose
& \\
\ours-FPN$^T$
& \reshl{35.4}{+}{2.4} & \reshl{63.3}{+}{3.5}  % coco pose
& \\
\end{tabular}
\end{raggedleft}
\caption{\textbf{Results on COCO dense pose estimation and Cityscapes instance segmentation tasks.} 
The performance deltas with the supervised baseline is shown in the brackets. \textcolor{Highlight}{\textbf{Green}}: deltas at least +0.5.}
\vspace{-2.0mm}
\label{tab:dp_and_cityscapes}
\end{table}

\paragraph{COCO DensePose Estimation.} 
We use DensePose R-CNN~\cite{guler2018densepose} with R50-FPN as our base model. As in COCO object detection and instance segmentation, we add Batch Normalization layers in FPN, and finetune and sync all such layers during training. We use the DensePose implementation in~\cite{wu2019detectron2}. The model is trained on~\texttt{train2014} split (1$\x$ schedule) and evaluated on~\texttt{val2014}. We report results under masked-GPS (AP$^{\text{gpsm}}$). Table~\ref{tab:dp_and_cityscapes} upper shows the comparisons of \ours versus MoCo-v2 and supervised pre-training counterparts. Our \ours-FPN improves over the supervised baseline by 2.2 points under \apdp{75}.

\paragraph{Cityscapes Instance Segmentation.} 
We use Mask R-CNN~\cite{he2017mask} with R50-FPN as our base model, add Batch Normalization layers before the FPN, and finetune and sync all such layers during training. The model is trained on the standard splits and training settings from~\cite{wu2019detectron2}. Table~\ref{tab:dp_and_cityscapes} lower shows the comparisons of \ours versus MoCo-v2 and supervised pre-training counterparts. All of the \ours variants substantially improve over the supervised baseline, \eg, by 3.3-3.6 points under AP$_{50}^{\text{mk}}$.

\section{Conclusion}
We presented \ours, a novel self-supervised representation learning algorithm for localization-based tasks such as object detection and segmentation. \ours performs both global and region-level contrastive learning to simultaneously learn semantic and spatial feature representations. The spatially consistent feature representations are learned on regions of the convolutional feature maps, while the global representations are learned over the entire image. In order to learn hierarchical representations, we further incorporated feature pyramids into the \ours pre-training, and showed how these parameters can also be transferred to popular FPN-based frameworks for downstream tasks.
Our method shows significant improvements for various localization-based tasks, such as object detection, instance segmentation, and dense pose estimation.

\paragraph{Acknowledgements}
Prof. Darrell’s group was supported in part by DoD including DARPA's XAI, LwLL, and SemaFor programs, as well as BAIR's industrial alliance programs. Prof. Keutzer's group was supported in part by Alibaba, Amazon, Google, Facebook, Intel, and Samsung as well as BAIR and BDD at Berkeley. Prof. Wang’s group was supported, in part, by gifts from Qualcomm and TuSimple. We would additionally like to thank Amir Bar, Ani Nrusimha, and Ilija Radosavovic for feedback and discussions. Wandb \cite{wandb} graciously provided academic research accounts to track our experimental analyses.

\newcommand{\appdx}{appendix}
\appendix

\section{Appendix}

\subsection{Experimentation Details}

\paragraph{Pre-training}
Table~\ref{tab:params} details the exact pre-training parameters used for \ours, where we followed the default training parameters from \cite{chen2020improved, chen2020exploring}. IN-100 experiments followed the exact same set of parameters except training occurred for 500 epochs with moco-k of 16,384 on IN-100, instead of 200 epochs and moco-k of 65,536 as was done on IN-1K (we used a smaller queue as IN-100 has fewer images).

\begin{table}[htb]
	\small
	\resizebox{1.0\linewidth}{!}{
\centering
\begin{tabular}{lll}
\toprule
\textbf{Parameter} & \textbf{MoCo-v2} & \textbf{SimSiam} \\
\midrule
 batch size & 256 & 256  \\
 num gpus & 8   & 8\\
 lr & 0.03 & 0.10 \\
 schedule & cosine & cosine\\
 opt & SGD  & SGD\\
 opt momentum & 0.9 & 0.9\\
 weight decay & 1e-4 & 1e-4\\
 epochs & 200  & 200 \\
 projection-mlp-dims & C5: $2048\rightarrow128$ & C5: $2048\rightarrow2048\rightarrow2048$ \\ 
 & C4: $1024\rightarrow128$ & C4: $1024\rightarrow1024\rightarrow1024$ 
 \\ & C3: $512\rightarrow128$ & C3: $512\rightarrow512\rightarrow512$
  \\
 moco-k & 65536 & - \\
 moco-m & 0.999 & - \\
 moco-t & 0.2 & - \\
 prediction-mlp-dims & - & C5: $2048\rightarrow512\rightarrow2048$ \\ 
 &  & C4: $1024\rightarrow256\rightarrow1024$ 
 \\ &  & C3: $512\rightarrow128\rightarrow512$\\
\bottomrule
\end{tabular}
}
 \vspace{0.1mm}
\caption{This table provides the parameters that were used for pre-training \ours-C4 and \ours-FPN for the MoCo-v2~\cite{chen2020improved} and SimSiam~\cite{chen2020exploring} implementations carried out in this paper (unless otherwise noted). The values in the brackets indicate a change in a parameter value when pretraining the SimSiam implementation compared to the MoCo-v2 version. IN-100 experiments followed the exact same set of parameters except training occurred for 500 epochs with moco-k of 16384 on IN-100, instead of 200 epochs and moco-k of 65536 as was done on IN-1K. The SimSiam projection and prediction dimensions indicate the dimensions of the MLP as specified in \cite{chen2020exploring}.}  
\label{tab:params}
\end{table}

\paragraph{Object detection}
The R50-C4 and R50-FPN backbones used for object detection are similar to those available in \texttt{Detectron2}~\cite{wu2019detectron2}, and followed the parameters settings and adjustments from~\cite{he2019momentum}. Specifically, for \emph{R50-C4}, the object detection backbone uses the output of the C4 stage, and the box prediction head uses the C5 stage with a batch norm layer following its output.

\paragraph{Selecting negative samples for \ours.} 
We use stride one average pooling of corresponding downsampling rates on feature maps of key views as negative samples. On $C4/P4$, since we use sliding window of size $48{\times}48$, where the feature map is downsampled 16x, the kernel size of average pooling is $3{\times}3$. The same rationale applies to $P3$ feature map. 

While MoCo uses a queue to maintain a large number of negative image-level samples, we find that such a queue is unnecessary for region-level samples as there are a large number of negative region samples within each batch. For similarity learning on $C4/P4$, we synchronize the pooled features across GPUs, leading to $12{\times}12{\times}256=36,864$ negative regions; for $P3$, we use the pooled features on each individual GPU, leading to $26{\times}26{\times}256/8=21,632$ negative samples. Note that we do not change the temperature hyperparameter in the contrastive learning objective, as the number of negatives for region-level similarity is roughly the same as the number of negatives for image-level similarity with momentum-based queue.

\subsection{Extended Experimental Results}

\paragraph{ImageNet linear probe performance}. While the object detection transfer performance leads to a substantial improvement compared to MoCo-v2, the \ours-C4 linear probe classification accuracy from ~\cite{chen2020improved} drops from $67.5\%$ to $66.1\%$ at 200 epochs of pre-training. This drop performance decrease indicates that ImageNet classification does not necessarily indicate an improved performance for region-level transfer tasks such as object detection. This observation was similarly reported by Chen et~al.~\cite{chen2020improved} where the authors observed that ``linear classification accuracy is not monotonically related to transfer performance in detection.''

% \clearpage

{\small
\bibliographystyle{ieee_fullname}
\bibliography{refs}

\begin{thebibliography}{10}\itemsep=-1pt

\bibitem{bachman2019learning}
Philip Bachman, R~Devon Hjelm, and William Buchwalter.
\newblock Learning representations by maximizing mutual information across
  views.
\newblock In {\em Advances in Neural Information Processing Systems},
  volume~32. Curran Associates, Inc., 2019.

\bibitem{bengio2013representation}
Yoshua Bengio, Aaron Courville, and Pascal Vincent.
\newblock Representation learning: A review and new perspectives.
\newblock {\em IEEE transactions on pattern analysis and machine intelligence},
  35(8):1798--1828, 2013.

\bibitem{wandb}
Lukas Biewald.
\newblock Experiment tracking with weights and biases, 2020.
\newblock Software available from wandb.com.

\bibitem{caron2020unsupervised}
Mathilde Caron, Ishan Misra, Julien Mairal, Priya Goyal, Piotr Bojanowski, and
  Armand Joulin.
\newblock Unsupervised learning of visual features by contrasting cluster
  assignments.
\newblock In {\em Thirty-fourth Conference on Neural Information Processing
  Systems (NeurIPS)}, 2020.

\bibitem{chen2020simple}
Ting Chen, Simon Kornblith, Mohammad Norouzi, and Geoffrey Hinton.
\newblock A simple framework for contrastive learning of visual
  representations.
\newblock In {\em International conference on machine learning}, pages
  1597--1607. PMLR, 2020.

\bibitem{chen2020improved}
Xinlei Chen, Haoqi Fan, Ross Girshick, and Kaiming He.
\newblock Improved baselines with momentum contrastive learning.
\newblock {\em arXiv preprint arXiv:2003.04297}, 2020.

\bibitem{chen2020exploring}
Xinlei Chen and Kaiming He.
\newblock Exploring simple siamese representation learning.
\newblock {\em arXiv preprint arXiv:2011.10566}, 2020.

\bibitem{cordts2016cityscapes}
Marius Cordts, Mohamed Omran, Sebastian Ramos, Timo Rehfeld, Markus Enzweiler,
  Rodrigo Benenson, Uwe Franke, Stefan Roth, and Bernt Schiele.
\newblock The cityscapes dataset for semantic urban scene understanding.
\newblock In {\em Proceedings of the IEEE conference on computer vision and
  pattern recognition}, pages 3213--3223, 2016.

\bibitem{deng2009imagenet}
Jia Deng, Wei Dong, Richard Socher, Li-Jia Li, Kai Li, and Li Fei-Fei.
\newblock Imagenet: A large-scale hierarchical image database.
\newblock In {\em 2009 IEEE conference on computer vision and pattern
  recognition}, pages 248--255. Ieee, 2009.

\bibitem{devlin2018bert}
Jacob Devlin, Ming-Wei Chang, Kenton Lee, and Kristina Toutanova.
\newblock Bert: Pre-training of deep bidirectional transformers for language
  understanding.
\newblock {\em arXiv preprint arXiv:1810.04805}, 2018.

\bibitem{ding2021unsupervised}
Jian Ding, Enze Xie, Hang Xu, Chenhan Jiang, Zhenguo Li, Ping Luo, and Gui-Song
  Xia.
\newblock Unsupervised pretraining for object detection by patch
  reidentification.
\newblock {\em arXiv preprint arXiv:2103.04814}, 2021.

\bibitem{donahue2014decaf}
Jeff Donahue, Yangqing Jia, Oriol Vinyals, Judy Hoffman, Ning Zhang, Eric
  Tzeng, and Trevor Darrell.
\newblock Decaf: A deep convolutional activation feature for generic visual
  recognition.
\newblock In {\em International conference on machine learning}, pages
  647--655, 2014.

\bibitem{dosovitskiy2015flownet}
Alexey Dosovitskiy, Philipp Fischer, Eddy Ilg, Philip Hausser, Caner Hazirbas,
  Vladimir Golkov, Patrick Van Der~Smagt, Daniel Cremers, and Thomas Brox.
\newblock Flownet: Learning optical flow with convolutional networks.
\newblock In {\em Proceedings of the IEEE international conference on computer
  vision}, pages 2758--2766, 2015.

\bibitem{dosovitskiy2014discriminative}
Alexey Dosovitskiy, Jost~Tobias Springenberg, Martin Riedmiller, and Thomas
  Brox.
\newblock Discriminative unsupervised feature learning with convolutional
  neural networks.
\newblock In {\em Advances in neural information processing systems}, pages
  766--774, 2014.

\bibitem{erhan2010does}
Dumitru Erhan, Yoshua Bengio, Aaron Courville, Pierre-Antoine Manzagol, Pascal
  Vincent, and Samy Bengio.
\newblock Why does unsupervised pre-training help deep learning?
\newblock {\em Journal of Machine Learning Research}, 11(Feb):625--660, 2010.

\bibitem{everingham2015pascal}
Mark Everingham, SM~Ali Eslami, Luc Van~Gool, Christopher~KI Williams, John
  Winn, and Andrew Zisserman.
\newblock The pascal visual object classes challenge: A retrospective.
\newblock {\em International journal of computer vision}, 111(1):98--136, 2015.

\bibitem{everingham2007pascal}
Mark Everingham, Luc Van~Gool, Christopher~KI Williams, John Winn, and Andrew
  Zisserman.
\newblock The pascal visual object classes challenge 2007 (voc2007) results.
\newblock 2007.

\bibitem{everingham2010pascal}
Mark Everingham, Luc Van~Gool, Christopher~KI Williams, John Winn, and Andrew
  Zisserman.
\newblock The pascal visual object classes (voc) challenge.
\newblock {\em International journal of computer vision}, 88(2):303--338, 2010.

\bibitem{gidaris2020learning}
Spyros Gidaris, Andrei Bursuc, Nikos Komodakis, Patrick P{\'e}rez, and Matthieu
  Cord.
\newblock Learning representations by predicting bags of visual words.
\newblock In {\em Proceedings of the IEEE/CVF Conference on Computer Vision and
  Pattern Recognition}, pages 6928--6938, 2020.

\bibitem{girshick2015fast}
Ross Girshick.
\newblock Fast r-cnn.
\newblock In {\em Proceedings of the IEEE international conference on computer
  vision}, pages 1440--1448, 2015.

\bibitem{goodfellow2016deep}
Ian Goodfellow, Yoshua Bengio, and Aaron Courville.
\newblock {\em Deep learning}.
\newblock MIT press, 2016.

\bibitem{goyal2019scaling}
Priya Goyal, Dhruv Mahajan, Abhinav Gupta, and Ishan Misra.
\newblock Scaling and benchmarking self-supervised visual representation
  learning.
\newblock In {\em Proceedings of the IEEE/CVF International Conference on
  Computer Vision}, pages 6391--6400, 2019.

\bibitem{grill2020bootstrap}
Jean-Bastien Grill, Florian Strub, Florent Altch{\'e}, Corentin Tallec,
  Pierre~H Richemond, Elena Buchatskaya, Carl Doersch, Bernardo~Avila Pires,
  Zhaohan~Daniel Guo, Mohammad~Gheshlaghi Azar, et~al.
\newblock Bootstrap your own latent: A new approach to self-supervised
  learning.
\newblock {\em arXiv preprint arXiv:2006.07733}, 2020.

\bibitem{guler2018densepose}
R{\i}za~Alp G{\"u}ler, Natalia Neverova, and Iasonas Kokkinos.
\newblock Densepose: Dense human pose estimation in the wild.
\newblock In {\em Proceedings of the IEEE conference on computer vision and
  pattern recognition}, pages 7297--7306, 2018.

\bibitem{he2019momentum}
Kaiming He, Haoqi Fan, Yuxin Wu, Saining Xie, and Ross Girshick.
\newblock Momentum contrast for unsupervised visual representation learning.
\newblock In {\em Proceedings of the IEEE Conference on Computer Vision and
  Pattern Recognition}, 2020.

\bibitem{he2017mask}
Kaiming He, Georgia Gkioxari, Piotr Doll{\'a}r, and Ross Girshick.
\newblock Mask r-cnn.
\newblock In {\em Proceedings of the IEEE international conference on computer
  vision}, pages 2961--2969, 2017.

\bibitem{he2016deep}
Kaiming He, Xiangyu Zhang, Shaoqing Ren, and Jian Sun.
\newblock Deep residual learning for image recognition.
\newblock In {\em Proceedings of the IEEE conference on computer vision and
  pattern recognition}, pages 770--778, 2016.

\bibitem{henaff2019data}
Olivier~J H{\'e}naff, Aravind Srinivas, Jeffrey De~Fauw, Ali Razavi, Carl
  Doersch, SM Eslami, and Aaron van~den Oord.
\newblock Data-efficient image recognition with contrastive predictive coding.
\newblock {\em arXiv preprint arXiv:1905.09272}, 2019.

\bibitem{jaiswal2021survey}
Ashish Jaiswal, Ashwin~Ramesh Babu, Mohammad~Zaki Zadeh, Debapriya Banerjee,
  and Fillia Makedon.
\newblock A survey on contrastive self-supervised learning.
\newblock {\em Technologies}, 9(1):2, 2021.

\bibitem{jiang2018acquisition}
Borui Jiang, Ruixuan Luo, Jiayuan Mao, Tete Xiao, and Yuning Jiang.
\newblock Acquisition of localization confidence for accurate object detection.
\newblock In {\em Proceedings of the European Conference on Computer Vision
  (ECCV)}, pages 784--799, 2018.

\bibitem{lecun2015deep}
Yann LeCun, Yoshua Bengio, and Geoffrey Hinton.
\newblock Deep learning.
\newblock {\em nature}, 521(7553):436--444, 2015.

\bibitem{li2019joint}
Xueting Li, Sifei Liu, Shalini De~Mello, Xiaolong Wang, Jan Kautz, and
  Ming-Hsuan Yang.
\newblock Joint-task self-supervised learning for temporal correspondence.
\newblock {\em arXiv preprint arXiv:1909.11895}, 2019.

\bibitem{lin2017feature}
Tsung-Yi Lin, Piotr Doll{\'a}r, Ross Girshick, Kaiming He, Bharath Hariharan,
  and Serge Belongie.
\newblock Feature pyramid networks for object detection.
\newblock In {\em Proceedings of the IEEE conference on computer vision and
  pattern recognition}, pages 2117--2125, 2017.

\bibitem{lin2014microsoft}
Tsung-Yi Lin, Michael Maire, Serge Belongie, James Hays, Pietro Perona, Deva
  Ramanan, Piotr Doll{\'a}r, and C~Lawrence Zitnick.
\newblock Microsoft coco: Common objects in context.
\newblock In {\em European conference on computer vision}, pages 740--755.
  Springer, 2014.

\bibitem{liu2020deep}
Li Liu, Wanli Ouyang, Xiaogang Wang, Paul Fieguth, Jie Chen, Xinwang Liu, and
  Matti Pietik{\"a}inen.
\newblock Deep learning for generic object detection: A survey.
\newblock {\em International journal of computer vision}, 128(2):261--318,
  2020.

\bibitem{loshchilov2016sgdr}
Ilya Loshchilov and Frank Hutter.
\newblock Sgdr: Stochastic gradient descent with warm restarts.
\newblock {\em arXiv preprint arXiv:1608.03983}, 2016.

\bibitem{malisiewicz2008recognition}
Tomasz Malisiewicz and Alexei~A Efros.
\newblock Recognition by association via learning per-exemplar distances.
\newblock In {\em 2008 IEEE Conference on Computer Vision and Pattern
  Recognition}, pages 1--8. IEEE, 2008.

\bibitem{minderer2019unsupervised}
Matthias Minderer, Chen Sun, Ruben Villegas, Forrester Cole, Kevin Murphy, and
  Honglak Lee.
\newblock Unsupervised learning of object structure and dynamics from videos.
\newblock {\em arXiv preprint arXiv:1906.07889}, 2019.

\bibitem{misra2020self}
Ishan Misra and Laurens van~der Maaten.
\newblock Self-supervised learning of pretext-invariant representations.
\newblock In {\em Proceedings of the IEEE/CVF Conference on Computer Vision and
  Pattern Recognition}, pages 6707--6717, 2020.

\bibitem{peng2018megdet}
Chao Peng, Tete Xiao, Zeming Li, Yuning Jiang, Xiangyu Zhang, Kai Jia, Gang Yu,
  and Jian Sun.
\newblock Megdet: A large mini-batch object detector.
\newblock In {\em Proceedings of the IEEE Conference on Computer Vision and
  Pattern Recognition}, pages 6181--6189, 2018.

\bibitem{pinheiro2020unsupervised}
Pedro~O Pinheiro, Amjad Almahairi, Ryan~Y Benmaleck, Florian Golemo, and Aaron
  Courville.
\newblock Unsupervised learning of dense visual representations.
\newblock {\em arXiv preprint arXiv:2011.05499}, 2020.

\bibitem{radford2018improving}
Alec Radford, Karthik Narasimhan, Tim Salimans, and Ilya Sutskever.
\newblock Improving language understanding by generative pre-training.

\bibitem{reed2020selfaugment}
Colorado~J Reed, Sean Metzger, Aravind Srinivas, Trevor Darrell, and Kurt
  Keutzer.
\newblock Selfaugment: Automatic augmentation policies for self-supervised
  learning.
\newblock In {\em Proceedings of the IEEE conference on Computer Vision and
  Pattern Recognition}, 2021.

\bibitem{ren2016faster}
Shaoqing Ren, Kaiming He, Ross Girshick, and Jian Sun.
\newblock Faster r-cnn: towards real-time object detection with region proposal
  networks.
\newblock {\em IEEE transactions on pattern analysis and machine intelligence},
  39(6):1137--1149, 2016.

\bibitem{thewlis2017unsupervised}
James Thewlis, Hakan Bilen, and Andrea Vedaldi.
\newblock Unsupervised learning of object frames by dense equivariant image
  labelling.
\newblock {\em arXiv preprint arXiv:1706.02932}, 2017.

\bibitem{tian2019contrastive}
Yonglong Tian, Dilip Krishnan, and Phillip Isola.
\newblock Contrastive multiview coding.
\newblock {\em arXiv preprint arXiv:1906.05849}, 2019.

\bibitem{tian2020makes}
Yonglong Tian, Chen Sun, Ben Poole, Dilip Krishnan, Cordelia Schmid, and
  Phillip Isola.
\newblock What makes for good views for contrastive learning?
\newblock {\em arXiv preprint arXiv:2005.10243}, 2020.

\bibitem{vondrick2018tracking}
Carl Vondrick, Abhinav Shrivastava, Alireza Fathi, Sergio Guadarrama, and Kevin
  Murphy.
\newblock Tracking emerges by colorizing videos.
\newblock In {\em Proceedings of the European conference on computer vision
  (ECCV)}, pages 391--408, 2018.

\bibitem{wang2019learning}
Xiaolong Wang, Allan Jabri, and Alexei~A Efros.
\newblock Learning correspondence from the cycle-consistency of time.
\newblock In {\em Proceedings of the IEEE Conference on Computer Vision and
  Pattern Recognition}, pages 2566--2576, 2019.

\bibitem{wang2020dense}
Xinlong Wang, Rufeng Zhang, Chunhua Shen, Tao Kong, and Lei Li.
\newblock Dense contrastive learning for self-supervised visual pre-training.
\newblock {\em arXiv preprint arXiv:2011.09157}, 2020.

\bibitem{wu2019detectron2}
Yuxin Wu, Alexander Kirillov, Francisco Massa, Wan-Yen Lo, and Ross Girshick.
\newblock Detectron2.
\newblock \url{https://github.com/facebookresearch/detectron2}, 2019.

\bibitem{wu2018unsupervised}
Zhirong Wu, Yuanjun Xiong, Stella~X Yu, and Dahua Lin.
\newblock Unsupervised feature learning via non-parametric instance
  discrimination.
\newblock In {\em Proceedings of the IEEE Conference on Computer Vision and
  Pattern Recognition}, pages 3733--3742, 2018.

\bibitem{xiao2020should}
Tete Xiao, Xiaolong Wang, Alexei~A Efros, and Trevor Darrell.
\newblock What should not be contrastive in contrastive learning.
\newblock In {\em International Conference on Learning Representations}, 2021.

\bibitem{xie2021detco}
Enze Xie, Jian Ding, Wenhai Wang, Xiaohang Zhan, Hang Xu, Zhenguo Li, and Ping
  Luo.
\newblock Detco: Unsupervised contrastive learning for object detection.
\newblock {\em arXiv preprint arXiv:2102.04803}, 2021.

\bibitem{yang2021instance}
Ceyuan Yang, Zhirong Wu, Bolei Zhou, and Stephen Lin.
\newblock Instance localization for self-supervised detection pretraining.
\newblock In {\em Proceedings of the IEEE/CVF Conference on Computer Vision and
  Pattern Recognition}, pages 3987--3996, 2021.

\bibitem{ye2019unsupervised}
Mang Ye, Xu Zhang, Pong~C Yuen, and Shih-Fu Chang.
\newblock Unsupervised embedding learning via invariant and spreading instance
  feature.
\newblock In {\em Proceedings of the IEEE/CVF Conference on Computer Vision and
  Pattern Recognition}, pages 6210--6219, 2019.

\bibitem{zeiler2014visualizing}
Matthew~D Zeiler and Rob Fergus.
\newblock Visualizing and understanding convolutional networks.
\newblock In {\em European conference on computer vision}, pages 818--833.
  Springer, 2014.

\bibitem{zhang2016colorful}
Richard Zhang, Phillip Isola, and Alexei~A Efros.
\newblock Colorful image colorization.
\newblock In {\em European conference on computer vision}, pages 649--666.
  Springer, 2016.

\end{thebibliography}
}

\end{document}